\pdfoutput=1
\documentclass[11pt]{article}

\usepackage[final]{ACL2023}

\usepackage{times}
\usepackage{latexsym}

\usepackage[T1]{fontenc}
\usepackage[utf8]{inputenc}

\usepackage{microtype}

\usepackage{inconsolata}

\usepackage{times}
\usepackage{latexsym}
\usepackage{danudefs}
\usepackage{algorithmic}
\usepackage{algorithm}
\usepackage{color}
\usepackage{booktabs}
\usepackage{xspace}
\usepackage{url}
\usepackage{arydshln}

\usepackage{acronym}
\acrodef{PoS}[PoS]{part-of-speech}
\acrodef{PIP}[PIP]{Pairwise Inner Product}
\acrodef{MLM}[MLM]{Masked Language Model}
\acrodef{SoTA}[SoTA]{state-of-the-art}
\usepackage[english]{babel}
\usepackage{hyperref}
\addto\captionsenglish{%
}
\addto\extrasenglish{%
}

\usepackage{xr}
\makeatletter
\newcommand*{\addFileDependency}[1]{% argument=file name and extension
  \typeout{(#1)}
  \@addtofilelist{#1}
  \IfFileExists{#1}{}{\typeout{No file #1.}}
}
\makeatother
\newcommand*{\myexternaldocument}[1]{%
    \externaldocument{#1}%
    \addFileDependency{#1.tex}%
    \addFileDependency{#1.aux}%
}
\myexternaldocument{Supplementary}
\usepackage{todonotes}

\makeatletter
\if@todonotes@disabled

\else

\fi
\makeatother

\title{\emph{Together We Make Sense}-- Learning Meta-Sense Embeddings from Pretrained Static Sense Embeddings}

\author{Haochen Luo$^\dagger$ \And
    Yi Zhou$^\diamondsuit$  \\
  University of Liverpool$^\dagger$, Cardiff University$^\diamondsuit$, Amazon$^\ddagger$\\
    {\tt haochen.luo@outlook.com} \\
  {\tt danushka@liverpool.ac.uk} \\
  {\tt zhouy131@cardiff.ac.uk}\And
  Danushka Bollegala$^{\dagger,\ddagger}$}

\date{}

\begin{document}
\maketitle

\begin{abstract}
Sense embedding learning methods learn multiple vectors for a given ambiguous word, corresponding to its different word senses.
For this purpose, different methods have been proposed in prior work on sense embedding learning that use different sense inventories, sense-tagged corpora and learning methods.
However, not all existing sense embeddings cover all senses of ambiguous words equally well due to the discrepancies in their training resources.
To address this problem, we propose the first-ever meta-sense embedding method -- Neighbour Preserving Meta-Sense Embeddings, which learns meta-sense embeddings by combining multiple independently trained source sense embeddings such that the sense neighbourhoods computed from the source embeddings are preserved in the meta-embedding space.
Our proposed method can combine source sense embeddings that cover different sets of word senses.
Experimental results on Word Sense Disambiguation (WSD) and Word-in-Context (WiC) tasks show that the proposed meta-sense embedding method consistently outperforms several competitive baselines.
\end{abstract}

\section{Introduction}
\label{sec:intro}

In contrast to static word embedding methods~\cite{MIklov-cbow-skip,pennington2014glove} that learn a vector, that represents the meaning of a word, sense embedding methods~\cite{LMMS,CardiffSense,scarlini-etal-2020-sensembert,ARES} learn multiple vectors per word, corresponding to the different senses of an ambiguous word.
Prior work has shown that sense embeddings are useful for tasks such as Word Sense Disambiguation (WSD) and sense discrimination tasks such as Word in Context (WiC)~\cite{Loureiro2019LIAADAS,Pilehvar:2019}.
However, existing sense embeddings are trained on diverse resources such as sense tagged corpora or dictionary glosses, with varying levels of sense coverage (e.g. fully-covering all synsets in the WordNet or only a subset), and using different methods.
Therefore, the performance reported by the existing sense embeddings on different downstream tasks and datasets vary significantly for different \ac{PoS} categories.
Moreover, it is not readily clear which sense embedding learning method should be used for disambiguating words in a given domain.

Meta-embedding learning has been successfully used to learn accurate and high coverage word- and sentence-level meta-embeddings by combining independently trained multiple source embeddings~\cite{Bollegala:IJCAIb:2022,yin2016learning}.
However, to the best of our knowledge, meta-embedding learning methods have \emph{not} been applied for sense embeddings before.
Compared to word-level meta-embedding, sense-level meta-embedding has two important challenges.
\paragraph{Challenge 1 (\emph{missing senses}).}
Compared to learning meta-word embeddings, where each word is assigned a single embedding, in static sense embeddings an ambiguous word is associated with multiple sense embeddings, each corresponding to a distinct sense of the ambiguous word.
However, not all of the different senses of a word might be equally covered by all source sense embeddings.
%How to handle the missing senses is an important problem when learning sense-level meta embeddings.
%Word-level meta embedding methods use different techniques to address this missing source embeddings problem such as assigning zero vectors for the missing word embeddings, or by first learning a transformation between each pair of source embeddings and then using the learnt transformation to predict the missing source embeddings.

\paragraph{Challenge 2 (\emph{Misalignment between sense and context embeddings}).}
In downstream tasks such as WSD, we must determine the correct sense $s$ of an ambiguous word $w$ in a given context (i.e. a sentence) $c$.
This is done by comparing the sense embeddings for each distinct sense of $w$ against the context embedding of $c$, for example, computed using a \ac{MLM} such as BERT~\cite{BERT}.
The sense corresponding to the sense embedding that has the maximum similarity with the context embedding is then selected as the correct sense of $w$ in $c$.
For sense embeddings such as LMMS~\cite{LMMS} or ARES~\cite{ARES} this is trivially achieved because they are both BERT-based embeddings and the cosine similarity between those sense embeddings and BERT embeddings can be directly computed.
However, this is \emph{not} the case for the meta-sense embeddings that exist in a different vector space than the context embeddings produced by BERT, where a projection between meta-sense and context embedding spaces must be learned before conducting WSD.

To address these challenges, we propose  \textbf{Neighbourhood Preserving Meta-Sense Embedding} (NPMS) by incorporating multiple independently trained \textbf{source} sense embeddings to learn a \textbf{meta}-sense embedding such that the sense-related information captured by the source (input) sense embeddings is preserved in the (output) meta-sense embedding.
NPMS can combine full-coverage sense embeddings with partial-coverage ones, thereby improving the sense coverage in the latter. 

NPMS does \emph{not} compare the source embeddings directly but require the nearest neighbours computed using source and meta sense embeddings to be similar.
We call this \emph{information preservation} criteria, and use \ac{PIP} to compare the similarity distributions (nearest neighbours) over senses between meta and source embedding spaces.
This allows us to address Challenge~1 using shared neighbours to compute the alignment between source and meta embedding spaces, without predicting any missing sense embeddings.
To address Challenge 2, NPMS requires meta-sense embedding of a word sense to be similar to the contextualised (word) embeddings of the words that co-occur in the same sentence. 
We call this \emph{contextual alignment}, and learn the sense-specific projection matrices that satisfy this criteria.
This ensures that meta-sense embeddings could be used in downstream tasks such as WSD or WiC, where we must select the correct sense of an ambiguous word given its context.

We evaluate NPMS on WiC and WSD  tasks against several competitive baselines for meta-embedding.
Experimental results show that NPMS consistently outperforms all other methods in both tasks.
More importantly, we obtain \ac{SoTA} performance for WSD and WiC tasks, reported by any static sense embedding method.
Source code for the proposed method is publicly available.\footnote{\url{https://github.com/LivNLP/NPMS}}

\section{Related Work}
\label{sec:related}

Our work is related to both static sense embeddings and meta-embedding learning as we review next.
\paragraph{Static Sense Embeddings} assign multiple embeddings for a single word, corresponding to its distinct senses.
\newcite{reisinger2010multi} proposed multi-prototype embeddings to represent word senses, which was extended by \newcite{huang2012improving} combining both local and global contexts.  
Both methods use a fixed number of clusters to represent a word, whereas \newcite{neelakantan2014efficient} proposed a non-parametric model, which estimates the number of senses dynamically per each word.
\newcite{chen-etal-2014-unified} initialised sense embeddings by means of glosses from WordNet, and adapted the skip-gram objective~\cite{mikolov} to learn and improve sense embeddings jointly with word embeddings.
\newcite{rothe-schutze-2015-autoextend} used pretrained word2vec embeddings to compose sense embeddings from sets of synonymous words.
\newcite{camacho2016nasari} created sense embeddings using structural knowledge from the BabelNet~\cite{navigli2010babelnet}.
\newcite{LMMS} constructed sense embeddings by taking the average over the contextualised embeddings of the sense annotated tokens from SemCor. 
\newcite{scarlini-etal-2020-sensembert} used the lexical-semantic information in BabelNet to produce sense embeddings without relying on sense-annotated data.
\newcite{ARES} also proposed ARES, a knowledge-based approach for constructing BERT-based embeddings of senses by means of the lexical-semantic information in BabelNet and Wikipedia.

\paragraph{Meta embedding learning} was first proposed for combining multiple pretrained static word embeddings~\cite{yin-schutze}.
Vector concatenation~\cite{Bollegala:IJCAIa:2022} is known to be a surprisingly strong baseline but increases the dimensionality of the meta-embedding with more sources.
\newcite{coates} showed that averaging performs comparable to concatenation under certain orthonormal conditions, while not increasing the dimensionality.
Learning orthogonal projections prior to averaging has shown to further improve performance~\cite{jawanpuria-etal-2020-learning}.
Globally linear~\cite{yin-schutze}, locally linear~\cite{bollegala2018learning} and autoencoder-based non-linear projections~\cite{autoencoder} have been used to learn word-level meta-embeddings.
Meta-embedding methods have been used for contextualised word embeddings~\cite{kiela-etal-2018-dynamic} and sentence embeddings~\cite{Takahashi:LREC:2022,Poerner:2020}.
For an extensive survey on meta-embedding learning see \newcite{Bollegala:IJCAIb:2022}.
However, to our best knowledge, we are the first to apply meta-embedding learning methods to learn sense embeddings.

\section{Meta-Sense Embedding Learning}
\label{sec:method}

To explain our proposed method in detail, let us first consider a vocabulary $\cV$ of words $w \in \cV$.
We further assume that each word $w$ is typically associated with one or more distinct senses $s$ and the set of senses associated with $w$ is denoted by $\cS_w$.
In meta-sense embedding learning, we assume a sense $s$ of a word to be represented by a set of $n$ source sense embeddings.
Let us denote the $j$-th source embedding of $s$ by $\vec{x}_j(s) \in \R^{d_{j}}$, where $d_j$ is the dimensionality of the $j$-th source embedding.

We project the $j$-th source embedding by a matrix $\mat{P}_j \in \R^{d \times d_{j}}$ into a common meta-sense embedding space with dimensionality $d$.
The meta-sense embedding, $\vec{m}(s) \in \R^d$ of $s$ is computed as the unweighted average of the projected source sense embeddings as given by \eqref{eq:ME}.
\begin{align}
    \label{eq:ME}
    \vec{m}(s) = \frac{1}{n} \sum_{j=1}^{n} \mat{P}_j\vec{x}_j(s)
\end{align}
After this projection step, all source sense embeddings live in the same $d$-dimensional vector space, thus enabling us to add them as done in \eqref{eq:ME}.

An advantage of considering the average of the projected source embeddings as the meta-sense embedding is that, even if a particular sense is not covered by one or more source sense embeddings, we can still compute a meta-sense embedding using the remainder of the source sense embeddings.
Moreover, prior work on word-level and sentence-level meta-embedding have shown that averaging after a linear projection improves performance when learning meta embeddings~\cite{coates,jawanpuria-etal-2020-learning,Poerner:2020}.

If we limit the projection matrices to be orthonormal, they can be seen as optimally rotating the source sense embeddings such that the projected source embeddings could be averaged in the meta-embedding space.
However, we observed that dropping this regularisation term to produce better meta-sense embeddings in our experiments.
Therefore, we did not impose any orthonormality restrictions on the projection matrices.

We require a meta-sense embedding to satisfy two criteria: (a) \textbf{sense information preservation} and (b) \textbf{contextual alignment}.
The two criteria jointly ensure that the meta-sense embeddings we learn are accurate and can be used in downstream tasks such as WSD in conjunction with contextualised word embeddings produced by an \ac{MLM}.
Next, we describe each of those criteria in detail.

\subsection{Sense Information Preservation}
\label{sec:preservation}

Given that the individual source sense embeddings are trained on diverse sense-related information sources, we would like to preserve this information as much as possible in the meta-sense embeddings we create from those source sense embeddings.
This is particularly important in meta-embedding learning because we might not have access to all the resources that were used to train the individual source sense embeddings, nor we will be training meta-embeddings from scratch but will be relying upon pretrained sense embeddings as the sole source of sense-related information into the meta-embedding learning process.
Therefore, we must preserve the complementary sense-related information encoded in the source sense embeddings as much as possible in their meta-sense embedding.

It is not possible however to directly compare the meta-sense embeddings computed using \eqref{eq:ME} against the source sense embeddings because they have different dimensionalities and live in different vector spaces.
This makes it challenging when quantifying the amount of information lost due to meta embedding using popular loss functions such as squared Euclidean distance between source and meta embeddings.
To address this problem we resort to \ac{PIP}, which has been previously used to determine the optimal dimensionality of word embeddings~\cite{Yin:2018} and learning concatenated word-level meta embeddings~\cite{Bollegala:IJCAIa:2022}.

Given a source/meta embedding matrix $\mat{E}$, the corresponding PIP matrix is given by \eqref{eq:PIP}
\begin{align}
    \label{eq:PIP}
    {\rm PIP}(\mat{E}) = \mat{E}\mat{E}\T
\end{align}

Specifically, PIP matrix contains the inner-products between all pairs of sense embeddings represented by the rows of $\mat{E}$.
${\rm PIP}(\mat{E})$ is a symmetric matrix with its number of rows (columns) equal to the total number of unique senses covering all the words in the vocabulary.
PIP matrices can be efficiently computed for larger dimensions and vocabularies because the inner-product computation can be parallelized over the embeddings.

Let us denote the source sense embedding matrix for the $j$-th source by $\mat{X}_j$, where the $i$-th row represents sense embedding $\vec{x}_j(s_i)$ learnt for the $i$-th sense $s_i$.
Likewise, let us denote by $\mat{M}$ the meta-sense embedding matrix, where the $i$-th row represents the meta-sense embedding $\vec{m}(s_i)$ computed for $s_i$ using \eqref{eq:ME}.
Because the shape of PIP matrices are independent from the dimensionalities of the embedding spaces, and the rows are aligned (i.e. sorted by the sense ids $s_i$), we can compare the meta-sense embedding against the individual source sense embedding using PIP loss, $L_{\rm pip}$, given by \eqref{eq:pip-loss}.
\begin{align}
    \label{eq:pip-loss}
    L_{\rm pip} = \sum_{j=1}^{n} \norm{\mathrm{PIP}(\mat{X}_j) - \mathrm{PIP}(\mat{M})}_F^2
\end{align}

Here, $\norm{\mat{A}}_F = \sqrt{\sum_{l,m}a^2_{lm}}$ denotes the Frobenius norm of the matrix $\mat{A}$.
PIP loss can be seen as comparing the distributions of similarity scores computed using the meta-sense embedding and each of the individual source sense embeddings for the same set of senses.
Although the actual vector spaces might be different and initially not well-aligned due to the projection and averaging steps in \eqref{eq:ME}, we would require the neighbourhoods computed for each word to be approximately similar in the meta-sense embedding space and each of the source sense embedding spaces.
PIP loss given in \eqref{eq:pip-loss} measures this level of agreement between meta and source embedding spaces.

\subsection{Contextual Alignment}
\label{sec:context}

The context in which an ambiguous word has been used provides useful clues to determine the correct sense of that word~\cite{yi-zhou-2021-learning}.
For example, consider the following two sentences: 
\textbf{(S1)} \emph{I went to the \textcolor{red}{bank} to withdraw some cash.}, and
\textbf{(S2)} \emph{The river \textcolor{blue}{bank} was crowded with people doing BBQs.}
Words \emph{cash} and \emph{withdraw} indicate the \emph{financial institute} sense of bank in \textbf{S1}, whereas the words \emph{river}, \emph{BBQ} indicate the \emph{sloping land} sense of bank in \textbf{S2}.

Let us denote the contextualised word embedding of a word $w$ in a context $c$ by $\vec{f}(w;c)$.
MLMs such as BERT and RoBERTa~\cite{RoBERTa} have been used in prior work in WSD to compute context-sensitive representations for ambiguous words.
Then, the above-described agreement between the sense $s$ of $w$ and its context $c$ can be measured by the similarity between the meta-sense embedding $\vec{m}(s)$ and the contextualised embedding $\vec{f}(w;c)$.
We refer to this requirement as the \emph{contextual alignment} between a meta-sense embedding and contextualised word embeddings.

Given a sense annotated dataset such as SemCor, we represent it by a set $\cT$ of tuples $(w,s,c)$, where the word $w$ is annotated with its correct sense $s$ in context $c$.
Then, we define the contextual alignment loss $L_{\rm cont}$ as (negative) average cosine similarity between $\vec{m}(s)$ and $\vec{f}(w;c)$, given by \eqref{eq:cont-loss}.
\begin{align}
    \label{eq:cont-loss}
    L_{\rm cont} = - \sum_{(w,s,c) \in \cT} \frac{\vec{m}(s)\T \vec{f}(w;c)}{\norm{\vec{m}(s)}_2 \norm{\vec{f}(w;c)}_2}
\end{align}

Minimising the contextual alignment loss in \eqref{eq:cont-loss}, will maximise the cosine similarity between the meta-sense embedding and the corresponding contextualised embedding.

In contrast to the PIP-loss defined by \eqref{eq:pip-loss}, which can be computed without requiring sense annotated data, the contextual alignment loss defined by \eqref{eq:cont-loss} requires sense annotated data.
However, SemCor, the sense annotated dataset that we use for computing the contextual alignment loss in this paper, is already being used by many existing pretrained source sense embeddings.
Therefore, we emphasise that we are \emph{not} requesting for any additional training resources during the meta-sense embedding learning process beyond what has been already used to train the source sense embeddings.
Moreover, ablation studies (\autoref{sec:results}) show that PIP-loss alone obtains significant improvements, without the contextual alignment loss.

Contextual alignment loss can also be motivated from an application perspective.
Sense embeddings are often used to represent word senses in downstream tasks such as WSD.
A typical approach for predicting the sense of an ambiguous word $w$ as used in a given context $c$ is to measure the cosine similarity between each sense embedding of $w$ and the context embedding for $c$~\cite{ARES,LMMS}.
The objective given in \eqref{eq:cont-loss} can be seen as enforcing this property directly into the meta-sense embedding learning process.
As we later see in \autoref{sec:exp}, NPMS perform particularly well in WSD benchmarks.

In order to be able to compute the cosine similarity between meta-sense embeddings and contextualised word embeddings, we must first ensure that they have the same dimensionality. 
This can be achieved by either (a) setting the dimensionality of the meta-sense embeddings equal to that of the contextualised word embeddings, or (b) by learning a projection matrix that adjusts the dimensionality of the meta-sense embeddings to that of the contextualised word embeddings.

\subsection{Parameter Learning}
\label{sec:learn}

We consider the linearly-weighted sum of the PIP-loss and contextual alignment loss as the total loss, $L_{\rm tot}$, given by \eqref{eq:total-loss}.
\begin{align}
    \label{eq:total-loss}
    L_{\rm tot}(\{\mat{P}_j\}_{j=1}^{n}) = \alpha L_{\rm pip} + (1 - \alpha) L_{\rm cont}
\end{align}
Here, the parameters to be learnt are the projection matrices $\mat{P}_j$ for the sources $j = 1,\ldots,n$.
The weighting coefficient $\alpha \in [0,1]$ is a hyperparameter determining the emphasis between the two losses.
In our experiments, we tune $\alpha$ using validation set of the Senseval-3 WSD dataset~\cite{snyder-palmer-2004-english}.

Compared to the cosine similarity, which is upper bounded by $1$, the PIP-loss grows with the size of the PIP matrices being used.
Therefore, we found that scaling the two losses by their mean values to be important to stabilise the training.
%If we want to ensure that the projection matrices remain orthonormal, we can include a regularisation term $\sum_{j=1}^{n} \norm{\mat{P}_j\mat{P}_j\T - \mat{I}}_F^2$ to \eqref{eq:total-loss}, where $\mat{I}$ is the identity matrix.
We initialise the projection matrices to the identity matrix and use vanilla stochastic gradient descent with a learning rate of $0.001$, determined using the validation set of the Senseval-3 WSD dataset.

\section{Experiments}
\label{sec:exp}

\subsection{Source Embeddings}
\label{sec:sources}

Our proposed NPMS is agnostic to the methods used to learn the source sense embeddings, and thus in principle can be used to meta-embed any source sense embedding.
In our experiments, we use the following source sense embeddings because of their accuracy, public availability, coverage word senses and diversity (i.e. trained on different resources to have different dimensionalities) such that we can conduct an extensive evaluation.

\paragraph{LMMS}\cite{LMMS} (\underline{L}anguage \underline{M}odelling \underline{M}akes \underline{S}ense) is a supervised approach to learn full-coverage static sense embeddings that cover all of the 206,949 senses in the WordNet.
We use three variants of LMMS~\cite{LMMS-reloaded} embeddings\footnote{\url{https://github.com/danlou/LMMS}} as sources in our experiments:
(a) \textbf{LMMS} (uses 1024 dimensional \texttt{bert-large-cased}\footnote{\url{https://huggingface.co/bert-large-cased}} embeddings with semantic networks (i.e., WordNet) and glosses to create 2048 dimensional sense emebddings),
%(b) \textbf{LMMS}$_{2348}$ (2348 dimensional sense embeddings created by appending the static word embedding generated from fastText~\cite{bojanowski-etal-2017-enriching} to increase robustness,
(b) \textbf{LMMS (XLNet)} (uses 1024 dimensional \texttt{xlnet-large-cased}\footnote{\url{https://huggingface.co/xlnet-large-cased}} as the base MLM, and averages contextualised embeddings computed from SemCor and WordNet glosses), and
(c) \textbf{LMMS (RoBERTa)} (uses 1024 dimensional \texttt{roberta-large}\footnote{\url{https://huggingface.co/roberta-large}} as the base MLM, and  and averages contextualised embeddings computed from SemCor and WordNet glosses).

\paragraph{SenseEmBERT}\cite{scarlini-etal-2020-sensembert} (\underline{Sense} \underline{Em}bedded \underline{BERT}) obviates the need for sense-annotated corpora by using the BabelNet\footnote{\url{babelnet.org}} mappings between WordNet senses and Wikipedia pages to construct sense embeddings with 2048 dimensions, covering all the 146,312 English nominal senses in the WordNet.
Each sense embedding consists of two components: (a) the average of the word embedding of the a target sense's relevant words, and (b) the average of the BERT encoded tokens of the sense gloss. 
For the brevity of the notation, we denote SenseEmBERT as \textbf{SBERT} in the remainder of this paper.

\paragraph{ARES}\cite{ARES} (context-\underline{A}wa\underline{R}e \underline{E}mbedding\underline{S}) is a semi-supervised method that learns sense embeddings with full-coverage of the WordNet and is 2048 dimensional.
ARES embeddings are created by applying BERT on the glossary information and the information contained in the SyntagNet \cite{maru-etal-2019-syntagnet}. 
It outperforms LMMS in WSD benchmarks.

\paragraph{DeConf}~\cite{deconf} are the 50-dimensional\footnote{\url{https://pilehvar.github.io/deconf/}} De-conflated Semantic Embeddings created from Wikipedia and Gigaword corpus using GloVe~\cite{pennington2014glove}.
DeConf enables us to evaluate the effect of combining a source that has significantly smaller dimensionality than the other source sense embeddings.

The intersection of the LMMS$_{2048}$ and ARES contains 206,949 senses, which is equivalent to the total number of senses in the WordNet because they both cover all the sense in the WordNet (i.e. full coverage sense embeddings). 
On the other hand, the intersection between the LMMS$_{2048}$ and SensEmBERT as well as the intersection between the ARES and SensEmBERT contains 146,312 senses, which is the total number of nominal senses in the WordNet.
By using source sense embeddings with different sense coverages we aim to evaluate the ability of meta-sense embedding methods to learn accurate sense embeddings by exploiting the complementary strengths in the sources.

\subsection{Evaluation Tasks}
\label{sec:eval-tasks}

We compare the accuracy of meta-sense embeddings using two standard tasks that have been used in prior work on sense embedding learning.

\paragraph{Word Sense Disambiguation (WSD):}
WSD is a longstanding problem in NLP, which aims to assign an ambiguous word in a context with a word sense~\cite{navigli2009word}.
To test whether NPMS can disambiguate the different senses of an ambiguous word, we conduct a WSD task using the evaluation framework proposed by~\newcite{raganato2017word}, which contains all-words English WSD datasets: Senseval-2~\cite[\textbf{SE2};][]{edmonds-cotton-2001-senseval}, Senseval-3~\cite[\textbf{SE3};][]{snyder-palmer-2004-english}, SemEval-07~\cite[\textbf{SE07};][]{pradhan-etal-2007-semeval}, SemEval-13~\cite[\textbf{SE13};][]{navigli-etal-2013-semeval} and SemEval-15~\cite[\textbf{SE15};][]{moro-navigli-2015-semeval}.
We use the official framework to avoid any discrepancies in the scoring methodology.

We perform WSD following the 1-NN procedure, where we compute the contextualised embedding, $\vec{f}(w;c)$, produced using an MLM.\footnote{In the case of BERT, we average the last four layers for each word $w$ in a test sentence $c$.}
We then measure the cosine similarity, $\phi(\vec{m}(s), \vec{f}(w;c))$, between the source/meta sense embedding for each sense $s$ of $w$, $\vec{m}(s)$, and $\vec{f}(w;c)$, and select the sense with the maximum cosine similarity as the correct sense of $w$ in $c$.

\paragraph{Word-in-Context (WiC):}
WiC is framed as a binary classification task, where given a target word $w$ and two contexts $c_2$ and $c_2$, the objective is to determine if $w$ occurring in $c_1$ and $c_2$ carries the same meaning.
A method that assigns the same vector to all senses of $w$ would report a chance-level (i.e. $50\%$) accuracy on WiC.

Given a target word $w$ in two contexts $c_1$ and $c_2$, we first determine the meta-sense embeddings of $w$, which are $\vec{m}(s_1)$ and $\vec{m}(s_2)$ corresponding to the senses of $w$ used in respectively $c_1$ and $c_2$. 
Let the contextualised word embedding of $w$ in $c_1$ and $c_2$ respectively be $\vec{f}(w;c_1)$ and $\vec{f}(w;c_2)$.
We train a binary logistic regression classifier on the WiC training set. 
Following the work from~\newcite{yi-zhou-2021-learning}, we use the cosine similarities between the two vectors in the following six pairs as features: $\phi(\vec{m}(s_1)$, $\vec{m}(s_2))$, $\phi(\vec{f}(w;c_{1})$, $\vec{f}(w;c_{2}))$, $\phi(\vec{m}(s_1)$, $\vec{f}(w;c_{1}))$, $\phi(\vec{m}(s_2)$, $\vec{f}(w;c_{2}))$, $\phi(\vec{m}(s_1)$, $\vec{f}(w;c_{2}))$ and $\phi(\vec{m}(s_2)$, $\vec{f}(w;c_{1}))$.

\subsection{Meta-Embedding Methods}
\label{sec:meta-embedding}

We extend prior works on word-level meta-embedding learning to meta-sense embedding learning by taking the sense embeddings described in \autoref{sec:sources} as source embeddings, and compare them with \textbf{NPMS} embeddings. 
We compare against the following methods:
\begin{itemize}
    \item \textbf{AVG}~\cite{coates} takes the average over the embeddings of a sense from different sources embeddings. 
    \item \textbf{CONC}~\cite{yin2016learning} creates meta-embeddings by concatenating the embeddings from different source embeddings.
    \item \textbf{SVD}~\cite{yin2016learning} performs dimensionality reduction on the concatenated source embeddings.
    \item \textbf{AEME}~\cite{bollegala2018learning} is an autoencoder-based method for meta-embedding learning, which is the current \ac{SoTA} unsupervised word-level meta-embedding learning method.
\end{itemize}
We use 2048 output dimensions for both SVD and AEME in the experiments, determined to be the best for those methods on validation data.

As noted in \autoref{sec:eval-tasks}, both WSD and WiC tasks require us to compute the cosine similarity, $\phi$, between a source/meta sense embedding, $\vec{m}(s)$, of a sense $s$ and a contextualised word embedding, $\vec{f}(w;c)$, of the ambiguous word $w$ in context $c$.
However, unlike for NPMS, which explicitly guarantees that its meta-sense embeddings are directly comparable with the contextualised word embeddings via the contextual loss \eqref{eq:cont-loss},
in general, the meta-sense embeddings produced by other methods do not always exist in the contextualised word embedding space associated with the MLM, which  requires careful consideration as discussed next.

As an concrete example, let us consider the meta-embedding of the three sources LMMS, ARES and SenseEmBERT, all of which are 2048 dimensional and computed by concatenating two 1024-dimensional BERT embeddings, averaged over different lexical resources.
Therefore, using the same 1024-dimensional BERT embeddings and by concatenating $\vec{f}(w;c)$ twice, we can obtain a 2048-dimensional BERT-based contextualised embedding for $w$ that can be used to compute the cosine similarity with a source sense embedding in this case.
We consider the meta-embedding of source sense embeddings with different dimensionalities and MLMs other than BERT such as LMMS (XLNet), LMMS (RoBERTa) and DeConf later in our experiments.

Next, let us consider the meta-sense embeddings produced by CONC.
Because the inner-product decomposes trivially over vector concatenation, we can copy and concatenate $\vec{f}(w;c)$ to match $\vec{m}(s)$ produced by CONC.
For example, if CONC is used with LMMS and ARES, we can concatenate $\vec{f}(w;c)$ four times, and then compute the inner-product with the meta-sense embedding.
AVG does not change the dimensionality of the meta-sense embedding space.
Therefore, we only need to concatenate $\vec{f}(w;c)$ twice when computing the cosine similarity with AVG for any number of source sense embeddings.

Unfortunately, the meta-sense embedding spaces produced by SVD and AEME are not directly comparable against that of contextualised embeddings due to the differences in dimensionality and non-linear transformations introduced (cf. AEME uses autoencoders).
Therefore, we learn a projection matrix, $\mat{A}$, between $\vec{m}(s)$ and $\vec{f}(w;c)$ by minimising the squared Euclidan distance given by \eqref{eq:squared-loss}, computed using the SemCor training dataset, $\cT$.
\begin{align}
    \label{eq:squared-loss}
    \sum_{(w,s,c) \in \cT} \norm{\mat{A}\vec{m}(s) - \vec{f}(w;c)}_2^2
\end{align}
After training, we compute the cosine similarity, $\phi(\mat{A}\vec{m}(s), \vec{f}(w;c))$, between the transformed SVD and AEME meta-sense embedding and contextualised embeddings.

\section{Results}
\label{sec:results}

\begin{table*}[t]\centering
%\resizebox{0.48\textwidth}{!}{
\begin{tabular}{lrrrrrrrr}\toprule
&SE2 & SE3 &SE07 &SE13 &SE15 &ALL &WiC \\ \midrule
LMMS &76.34 &75.57 &68.13 &75.12 &77.01 &75.44 &69.30 \\
ARES &78.05 &77.08 &70.99 &77.31 & \textbf{83.17} &77.91 &68.50 \\
SBERT &53.11 &52.22 &41.37 & \textbf{78.77} &55.12 &59.85 &71.14 \\ \midrule
AVG &79.36 & \textbf{77.46} &70.33 &77.86 &80.82 &78.17 &71.16 \\
CONC &78.22 &77.14 &70.99 &77.37 &82.97 &77.97 &70.38 \\
SVD &75.02 &74.22 &67.25 &72.81 &74.85 &73.80 &63.01 \\
AEME &78.53 &76.92 &69.01 &76.09 &78.96 &77.03 &70.69 \\
NPMS & \textbf{79.93} &77.30 & \textbf{71.65} &77.49 &81.21 & \textbf{78.37} & \textbf{71.47} \\
\bottomrule
\end{tabular}
%}
\caption{F1 scores on WSD benchmarks and accuracy on WiC are shown for the three sources (top) and for the different meta-embedding methods (bottom).}
\label{tbl:three-sources}
\end{table*}

\begin{table*}[t]\centering
%\resizebox{0.49\textwidth}{!}{
\begin{tabular}{lrrrrrrr}\toprule
&SE2 & SE3 &SE07 &SE13 &SE15 &ALL & WiC\\\midrule
\multicolumn{8}{c}{LMMS(BERT) [2048] + ARES (BERT) [2048]} \\
AVG &\textbf{78.79} &77.03 &69.89 &77.13 & \textbf{81.80} &77.83 &70.22 \\
NPMS &78.53 &\textbf{77.14} &\textbf{71.87} &\textbf{77.37} &81.60 &\textbf{77.93} &70.22 \\ \midrule

\multicolumn{8}{c}{ARES (BERT) [2048] + SBERT [2048]}\\
AVG &78.57 &77.35 &71.21 &78.10 &\textbf{81.70} &78.13 &71.32 \\
NPMS &\textbf{78.79} &77.41 &\textbf{71.65} &\textbf{78.53} &81.41 &\textbf{78.30} &71.32 \\ \midrule

\multicolumn{8}{c}{LMMS (BERT) [2048] + SBERT [2048]}\\
AVG &77.70 &76.16 &68.79 &78.04 &77.69 &76.82 &69.59 \\
NPMS &\textbf{78.05}	&\textbf{76.86}	&\textbf{69.89}	&\textbf{78.28}	&\textbf{78.28}	&\textbf{77.32} &\textbf{71.79} \\ \midrule

\multicolumn{8}{c}{LMMS(XLNet) [1024] + DeConf [50]} \\
AVG &40.80 &35.68 &21.32 &41.61 &43.93 &38.89 & 66.46\\
NPMS & \textbf{50.88} & \textbf{41.68} & \textbf{40.66} & \textbf{53.04} & \textbf{53.13} &\textbf{48.70} & \textbf{69.26}\\  \midrule

\multicolumn{8}{c}{LMMS(RoBERTa) [1024] + DeConf [50]} \\
AVG  &39.35 &34.97 &26.15 &41.48 &42.47 &38.33 & 66.46\\
NPMS & \textbf{48.77} & \textbf{44.81} & \textbf{39.34} & \textbf{53.41} & \textbf{53.52} &\textbf{48.89} & \textbf{69.75}\\  \bottomrule
\end{tabular}
%}
\caption{Meta-sense embedding of sources with different dimensionalities (shown in brackets) and MLMs.}
\label{tbl:diff}
\end{table*}

\subsection{Effect of Meta-embedding Learning}
 \autoref{tbl:three-sources} compares the performance of NPMS against the meta-embedding methods described in \autoref{sec:meta-embedding} on WSD and WiC.
We see that NPMS obtains the overall best performance for WSD (ALL) as well as on WiC.
Among the three sources, ARES reports the best performance for WSD (ALL), while SBERT does so for WiC.
In SE2, SE07 datasets NPMS reports the best performance, whereas AVG, SBERT and ARES do so respectively in SE3, SE13 and SE15.
Among the baseline methods, we see AVG to report the best results, which is closely followed by CONC.
Poor performance of SVD shows the challenge of applying dimensionality reduction methods on CONC due to missing sense embeddings.
Although AEME has reported the \ac{SoTA} performance for word-level meta-embedding, applying it directly on sense embeddings is suboptimal.
This shows the difference between word- vs. sense-level meta-embedding learning problems, and calls for sense-specific meta-embedding learning methods.

According to the WiC leader board,\footnote{\url{https://pilehvar.github.io/wic/}} the performance reported by NPMS is second only to SenseBERT~\cite{levine-etal-2020-sensebert}, which is a contextualised sense embedding method obtained by fine tuning BERT on WordNet supersenses.
Therefore, the performance of NPMS can be seen as the \ac{SoTA} for any \emph{static} sense embedding method.

\subsection{Effect of Source Embeddings}

The performance of a meta-embedding depends on the source embeddings used.
Therefore, we evaluate the ability of NPMS to create meta-sense embeddings from diverse source sense embeddings that have different dimensionalities and created from different MLMs.
Due space limitations, in \autoref{tbl:diff} we compare NPMS against AVG, which reported the best performance among all other meta-embedding learning methods in \autoref{tbl:three-sources}.
From \autoref{tbl:diff}, we see that when the dimensionalities of the two source sense embeddings are identical (i.e. 2048 dimensional LMMS + ARES or LMMS + SBERT configurations) or similar (i.e. 2048 dimensional ARES + 2048 dimensional SBERT configuration), AVG closely matches the performance of NPMS in WSD and WiC evaluations.
However, we see a drastically different trend when the two sources are not BERT-based (e.g. XLNet, RoBERTa) or when they have significantly different dimensionalities (1024 dimensional LMMS (XLNet), LMMS (RoBERTa) and 50 dimensional DeConf).
In such settings, we see that NPMS to performs significantly better than AVG across all WSD benchmarks as well as on WiC.
Recall that AVG assumes (a) the source embedding spaces to be orthogonal,
and (b) applies zero-padding to the smaller dimensional source embeddings to make them aligned with the rest of the source embeddings.
Both of those assumptions do not hold true when the source embeddings are created from diverse MLMs or have significantly different numbers of dimensions, which leads to suboptimal performances in AVG.
On the other hand, NPMS does \emph{not} directly compare source sense embeddings, but instead consider neighbourhoods computed from the source sense embeddings.
Moreover, zero-padding is not required in NPMS because the contextual alignment step ensures the proper alignment between the contextual embedding and meta-sense embedding spaces.
These advantages of NPMS are clearly evident from \autoref{tbl:diff}.

\subsection{Effect of Projection Learning}

\begin{table}[t]
\centering
\begin{tabular}{l c c}\toprule
     Method & WSD (ALL) & WiC  \\ \midrule
     SVD with proj. & 74.80 & 66.93 \\
     SVD without proj.& 35.90 & 60.34 \\ \midrule
     AEME with proj. & 76.02 & 68.65 \\
     AEME without proj. & 41.60 & 53.61 \\ \bottomrule
\end{tabular}
\caption{Effect of learning a projection matrix between meta-sense vs. BERT embedding spaces.}
\label{tbl:proj}
\end{table}

\begin{table*}[t]
\centering
%\resizebox{0.48\textwidth}{!}{
\begin{tabular}{lrrrrrrrr}\toprule
&SE2 & SE3 &SE07 &SE13 &SE15 &ALL &WiC \\\cmidrule{1-8}
%\multicolumn{8}{c}{ARES+LMMS}  \\ 
%NPMS &78.53 &\textbf{77.14} &\textbf{71.87} &77.37 &\textbf{81.60} &\textbf{77.93} &70.22 \\
%PIP &78.66 &76.76 &71.65 &\textbf{77.43} &81.51 &77.86 &70.69 \\
%CONT &\textbf{79.14} &77.03 &70.11 &76.95 &80.82 &77.77 &69.59 \\\cmidrule{1-8}

%\multicolumn{8}{c}{ARES+SBERT} \\
%NPMS &78.79 &\textbf{77.41} &71.65 &78.53 &81.41 &78.30 &71.32 \\
%PIP &78.70 &\textbf{77.41} &71.43 &\textbf{78.65} &\textbf{81.70} &\textbf{78.33} &71.94 \\
%CONT &\textbf{79.14} &\textbf{77.41} &\textbf{72.09} &78.28 &80.92 &78.31 &71.63 \\\cmidrule{1-8}

%\multicolumn{8}{c}{LMMS+SBERT} \\
%NPMS &78.09 &75.78 &70.11 &\textbf{78.04} &\textbf{78.08} &76.99 &71.63 \\
%PIP &77.87 &76.22 &68.13 &77.92 &77.69 &76.82 &71.79 \\
%CONT &\textbf{78.26} &\textbf{76.86} &\textbf{70.55} &77.49 &77.89 &\textbf{77.20} &72.57 \\\cmidrule{1-8}

%\multicolumn{8}{c}{ARES+LMMS+SBERT} \\
Both &\textbf{79.93}	&\textbf{77.30}	&71.65	&77.49	&\textbf{81.21}	&\textbf{78.37}	& \textbf{71.47} \\
$L_{\rm pip}$ only & 79.80 &77.03 &\textbf{71.87} &77.49 &80.72 &78.20 &70.69\\
$L_{\rm cont}$ only &79.54 &77.19 &70.77 &\textbf{77.86} &80.33 &78.12 &71.32 \\
\cmidrule{1-8}
\end{tabular}
%}
\caption{Ablation between the PIP-loss ($L_{\rm pip}$) and contextual alignment loss ($L_{\rm cont}$).}
\label{tbl:ablation}
\end{table*}

\autoref{tbl:proj} shows the importance of learning a projection matrix via \eqref{eq:squared-loss} between meta-sense and contextualised embeddings, for SVD and AEME.
We see that the performance of both of those methods drop significantly without the projection matrix learning step.
Even with projection matrices, SVD and AEME do not outperform simpler baselines such as AVG or CONC.
On the other hand, NPMS does not require such a projection matrix learning step and consistently outperforms all those methods across multiple WSD and WiC benchmarks.

\subsection{Effect of the Two losses}
To understand the contributions of the two loss terms PIP-loss ($L_{\rm pip}$) and contextual alignment loss ($L_{\rm cont}$), we conduct an ablation study where we train NPMS with three sources using only one of the two losses at a time.
From \autoref{tbl:ablation}, we see that in both WiC and WSD (ALL, SE2, SE3, SE15), the best performance is obtained by using both losses.
Each loss contributes differently in different datasets, although the overall difference between the two losses is non-significant (according to a paired Student's $t$-test with $p < 0.05$). 
This is particularly encouraging because PIP-loss can be computed without having access to a sense labelled corpus such as SemCor.
Such resources might not be available in specialised domains such as medical or legal texts.
Therefore, in such cases we can still apply NPMS trained using only the PIP-loss.
Although we considered a linearly-weighted combination of the two losses in \eqref{eq:total-loss}, we believe further improvements might be possible by exploring more complex (nonlinear) combinations of the two losses.
However, exploring such combinations is beyond the scope of current paper and is deferred to future work.
 
% NPMS is benefited from two loss terms when meta-sense embeddings are learnt with all the three source embeddings or the combination of ARES and LMMS. 
%For ARES+SBERT and LMMS+SBERT, NPMS does not show a significant improvement by using these two losses together.
%This implies that more advanced ways to combine these two loss terms may be done and NPMS method can achieve a even better performance.

\section{Conclusion}
\label{sec:conclusion}

We proposed the first-ever meta-sense embedding learning method.
Experimental results on WiC and WSD datasets show that our proposed NPMS surpasses previously published results for static sense embedding, and outperforms multiple word-level meta-embedding learning methods when applied to sense embeddings.
Our evaluations were limited to English and we will consider non-English sense embeddings in our future work.

\section{Limitations}

All source sense embeddings we used in our experiments are only covering the English language, which is morphologically limited.
Therefore, it is unclear whether our results and conclusions will still be valid for meta-sense embeddings created for languages other than English.
On the other hand, there are WSD and WiC benchmarks for other languages such as SemEval-13, SemEval-15, XL-WSD~\cite{pasini2021xl} and  WiC-XL~\cite{raganato-etal-2020-xl}, as well as multilingual sense embeddings such as ARES$_m$~\cite{ARES} and SensEmBERT~\cite{scarlini-etal-2020-sensembert}.
Extending our evaluations to cover multilingual sense embeddings is deferred to future work.

Our meta-sense embedding method requires static sense embeddings, and cannot be applied to contextualised sense embedding methods such as SenseBERT~\cite{levine-etal-2020-sensebert}.
There have been some work on learning word-level and sentence-level~\cite{Takahashi:LREC:2022,Poerner:2020} meta-embeddings using contextualised word embeddings produced by MLMs as the source embeddings.
However, contextualised sense embedding methods are limited compared to the numerous static sense embedding methods.
This is partly due to the lack of large-scale sense annotated corpora, required to train or fine-tune contextualised sense embeddings.
Extending our work to learn meta-sense embeddings using contextualised word embeddings as source embeddings is an interesting future research direction.

\section{Ethical Considerations}

We compared our proposed method, NPMS, with several baselines on WSD and WiC tasks.
In this work, we did not annotate any datasets by ourselves and used corpora and benchmark datasets that have been collected, annotated and repeatedly used for evaluations in prior works.
To the best of our knowledge, no ethical issues have been reported concerning these datasets.
Nevertheless, prior work from~\citet{zhou-etal-2022-sense} shows that pretrained sense embeddings encode various types of social biases such as gender and racial biases.
Moreover, it has also been reported recently that word-level meta-embedding methods can amplify the social biases encoded in the source embeddings~\cite{Kaneko:EMNLP:2022}.
Therefore, we emphasise that it is important to evaluate the meta-sense embeddings learnt in this work for unfair social biases before they are deployed to downstream applications.

\section*{Acknowledgements}
Danushka Bollegala holds concurrent appointments as a Professor at University of Liverpool and as an Amazon Scholar. This paper describes work performed at the University of Liverpool and is not associated with Amazon.

\bibliography{myrefs.bib}
\bibliographystyle{acl_natbib}

\end{document}